\def\year{2019}\relax
\documentclass[letterpaper]{article} 
\usepackage{aaai19}  
\usepackage{times}  
\usepackage{helvet}  
\usepackage{courier}  
\usepackage{url}  
\usepackage{graphicx}  
\usepackage{amsmath}
\usepackage{amssymb}
\usepackage{multirow}
\usepackage{subfigure}
\begin{document}
%
\title{Graph Convolutional Networks for Text Classification}
\author{Liang Yao, Chengsheng Mao, Yuan Luo\thanks{Corresponding Author}\\
Northwestern University\\
Chicago IL 60611\\
\{liang.yao, chengsheng.mao, yuan.luo\}@northwestern.edu\\
}
\maketitle
\begin{abstract}  
Text classification is an important and classical problem in natural language processing. There have been a number of studies that applied convolutional neural networks (convolution on regular grid, e.g., sequence) to classification. However, only a limited number of studies have explored the more flexible graph convolutional neural networks (convolution on non-grid, e.g., arbitrary graph) for the task. In this work, we propose to use graph convolutional networks for text classification. We build a single text graph for a corpus based on word co-occurrence and document word relations, then learn a Text Graph Convolutional Network (Text GCN) for the corpus. Our Text GCN is initialized with one-hot representation for word and document, it then jointly learns the embeddings for both words and documents, as supervised by the known class labels for documents. Our experimental results on multiple benchmark datasets demonstrate that a vanilla Text GCN without any external word embeddings or knowledge outperforms state-of-the-art methods for text classification. On the other hand, Text GCN also learns predictive word and document embeddings. In addition, experimental results show that the improvement of Text GCN over state-of-the-art comparison methods become more prominent as we lower the percentage of training data, suggesting the robustness of Text GCN to less training data in text classification.
%
%
\end{abstract}

\section{Introduction}
Text classification is a fundamental problem in natural language processing (NLP). There are numerous applications of text classification such as document organization, news filtering, spam detection, opinion mining, and computational phenotyping~\cite{aggarwal2012survey,zeng2018natural}. An essential intermediate step for text classification is text representation. Traditional methods represent text with hand-crafted features, such as sparse lexical features (e.g., bag-of-words and n-grams). Recently, deep learning models have been widely used to learn text representations, including convolutional neural networks (CNN) ~\cite{kim2014convolutional} and recurrent neural networks (RNN) such as long short-term memory (LSTM) ~\cite{hochreiter1997long}. As CNN and RNN prioritize locality and sequentiality~\cite{battaglia2018relational}, these deep learning models can capture semantic and syntactic information in local consecutive word sequences well, but may ignore global word co-occurrence in a corpus which carries non-consecutive and long-distance semantics~\cite{peng2018large}.
%
%

Recently, a new research direction called graph neural networks or graph embeddings has attracted wide attention~\cite{battaglia2018relational,cai2018comprehensive}. Graph neural networks have been effective at tasks thought to have rich relational structure and can preserve global structure information of a graph in graph embeddings.

In this work, we propose a new graph neural network-based method for text classification. We construct a single large graph from an entire corpus, which contains words and documents as nodes. We model the graph with a Graph Convolutional Network (GCN)~\cite{kipf2017semi}, a simple and effective graph neural network that captures high order neighborhoods information. The edge between two word nodes is built by word co-occurrence information and the edge between a word node and document node is built using word frequency and word's document frequency. We then turn text classification problem into a node classification problem. The method can achieve strong classification performances with a small proportion of labeled documents and learn interpretable word and document node embeddings. Our source code is available at~\url{https://github.com/yao8839836/text_gcn}. To summarize, our contributions are as follows:
\begin{itemize}
\setlength\itemsep{0em}
\item We propose a novel graph neural network method for text classification. To the best of our knowledge, this is the first study to model a whole corpus as a heterogeneous graph and learn word and document embeddings with graph neural networks jointly. 
\item Results on several benchmark datasets demonstrate that our method outperforms state-of-the-art text classification methods, without using pre-trained word embeddings or external knowledge. Our method also learn predictive word and document embeddings automatically.
\end{itemize}

\section{Related Work}
\subsection{Traditional Text Classification}
Traditional text classification studies mainly focus on feature engineering and classification algorithms. For feature engineering, the most commonly used feature is the bag-of-words feature. In addition, some more complex features have been designed, such as n-grams~\cite{wang2012baselines} and entities in ontologies~\cite{chenthamarakshan2011concept}. There are also existing studies on converting texts to graphs and perform feature engineering on graphs and subgraphs~\cite{luo2016bridging,rousseau2015text,skianis2016regularizing,luo2014automatic,luo2015subgraph}.  Unlike these methods, our method can learn text representations as node embeddings automatically.
%
%
%
%

\subsection{Deep Learning for Text Classification}
Deep learning text classification studies can be categorized into two groups.
One group of studies focused on models based on word embeddings~\cite{mikolov2013distributed,pennington2014glove}. Several recent studies showed that the success of deep learning on text classification largely depends on the effectiveness of the word embeddings~\cite{Shen2018Baseline,joulin2017bag,P18-1216}. Some authors aggregated unsupervised word embeddings as document embeddings then fed these document embeddings into a classifier \cite{le2014distributed,joulin2017bag}. Others jointly learned word/document and document label embeddings~\cite{tang2015pte,P18-1216}. Our work is connected to these methods, the major difference is that these methods build text representations after learning word embeddings while we learn word and document embeddings simultaneously for text classification.

Another group of studies employed deep neural networks. Two representative deep networks are CNN and RNN. \cite{kim2014convolutional} used CNN for sentence classification. The architecture is a direct application of CNNs as used in computer vision but with one dimensional convolutions. \cite{zhang2015character} and \cite{conneau2017very} designed character level CNNs and achieved promising results. 
\cite{tai2015improved}, \cite{liu2016recurrent} and \cite{luo2017recurrent} used LSTM, a specific type of RNN, to learn text representation. To further increase the representation flexibility of such models, attention mechanisms have been introduced as an integral part of models employed for text classification~\cite{yang2016hierarchical,wang2016attention}. 
%
%
Although these methods are effective and widely used, they mainly focus on local consecutive word sequences, but do not explicitly use global word co-occurrence information in a corpus.

\subsection{Graph Neural Networks}

The topic of Graph Neural Networks has received growing attentions recently~\cite{cai2018comprehensive,battaglia2018relational}. A number of authors generalized well-established neural network models like CNN that apply to regular grid structure (2-d mesh or 1-d sequence) to work on arbitrarily structured graphs~\cite{bruna2013spectral,henaff2015deep,defferrard2016convolutional,kipf2017semi}.
In their pioneering work, Kipf and Welling presented a simplified graph neural network model, called graph convolutional
networks (GCN), which achieved state-of-the-art classification results on a number of benchmark graph datasets ~\cite{kipf2017semi}.
GCN was also explored in several NLP tasks such as semantic role labeling~\cite{marcheggiani2017encoding}, relation classification~\cite{li2018gcrn} and machine translation~\cite{bastings2017graph}, where GCN is used to encode syntactic structure of sentences.
Some recent studies explored graph neural networks for text classification~\cite{henaff2015deep,defferrard2016convolutional,kipf2017semi,peng2018large,DBLP:conf/acl/ZhangLS18}. 
However, they either viewed a document or a sentence as a graph of word nodes~\cite{defferrard2016convolutional,peng2018large,DBLP:conf/acl/ZhangLS18} or relied on the not-routinely-available document citation relation to construct the graph~\cite{kipf2017semi}. In contrast, when constructing the corpus graph, we regard the documents and words as nodes (hence heterogeneous graph) and do not require inter-document relations.

\begin{figure*}[t]
  \centering
  \includegraphics[height=50 mm]{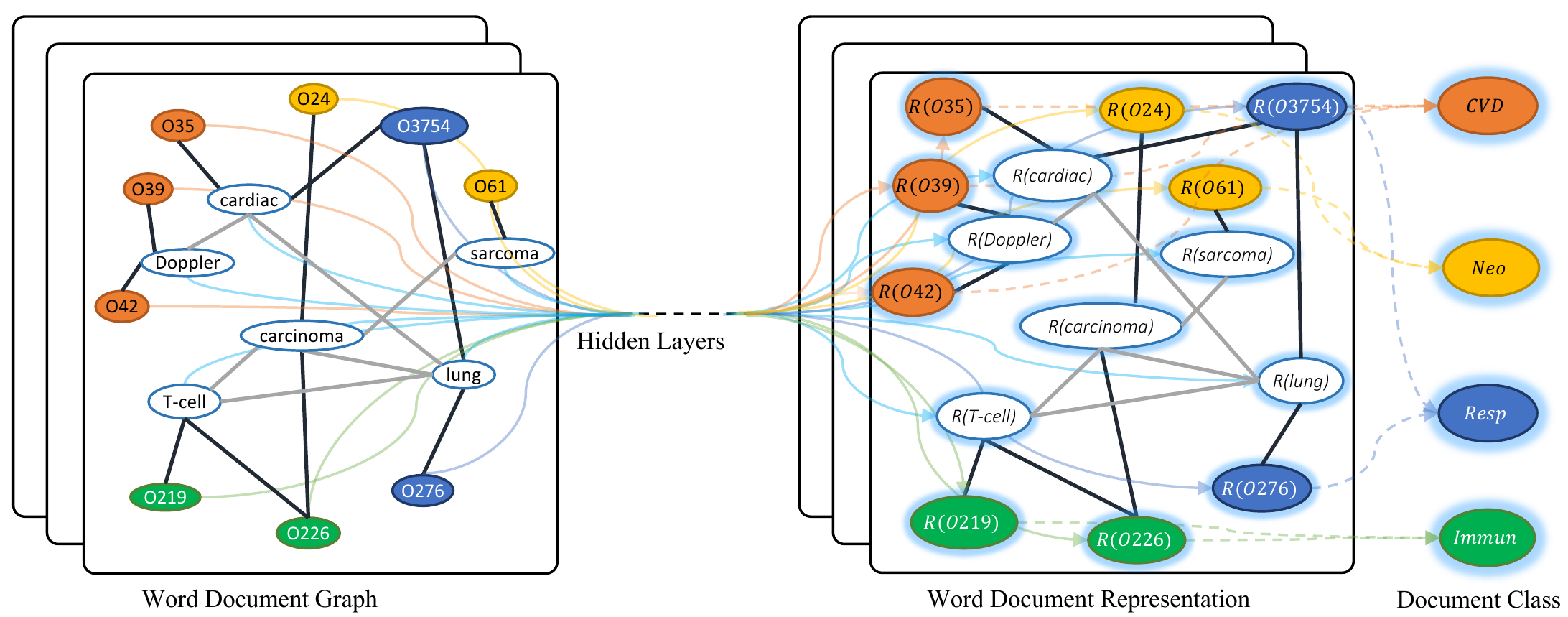}
  \caption{Schematic of Text GCN. Example taken from Ohsumed corpus. Nodes begin with ``O" are document nodes, others are word nodes. Black bold edges are document-word edges and gray thin edges are word-word edges. $R(x)$ means the representation (embedding) of $x$. Different colors mean different document classes (only four example classes are shown to avoid clutter). CVD: Cardiovascular Diseases, Neo: Neoplasms, Resp: Respiratory Tract Diseases, Immun: Immunologic Diseases.}
  \label{fig:framework}
\end{figure*}

\section{Method}
\subsection{Graph Convolutional Networks (GCN)}

A GCN~\cite{kipf2017semi} is a multilayer neural network that operates directly on a graph and induces embedding vectors of nodes based on properties of their neighborhoods. Formally, consider a graph $G = (V, E)$, where $V(|V| = n)$ and $E$ are sets of nodes and edges, respectively. Every node is assumed to be connected to itself, i.e., $(v, v) \in E$ for any $v$. Let $X \in \mathbb{R}^{n \times m}$ be a matrix containing all $n$ nodes with their features, where $m$ is the dimension of the feature vectors, each row $x_v \in \mathbb{R}^m$ is the feature vector for $v$. 
We introduce an adjacency matrix $A$ of $G$ and its degree matrix $D$, where $D_{ii} = \sum_j A_{ij}$. The diagonal elements of $A$ are set to $1$ because of self-loops.
GCN can capture information only about immediate neighbors with one layer of convolution. When multiple GCN layers are stacked, information about larger neighborhoods are integrated. For a one-layer GCN, the new $k$-dimensional node feature matrix $L^{(1)} \in \mathbb{R}^{n \times k}$ is computed as
\begin{equation}
L^{(1)} = \rho(\tilde{A}XW_0)
\label{1stl}
\end{equation}
where $\tilde{A} = D^{-\frac{1}{2}}A D^{-\frac{1}{2}}$ is the normalized symmetric adjacency matrix and $W_0 \in \mathbb{R}^{m \times k}$ is a weight matrix. $\rho$ is an activation function, e.g. a ReLU $\rho(x) = \text{max}(0,x)$. As mentioned before, one can incorporate higher order neighborhoods information by stacking multiple GCN layers:
\begin{equation}
L^{(j+1)} = \rho(\tilde{A}L^{(j)}W_j)
\end{equation}
where $j$ denotes the layer number and $L^{(0)} = X$.

%
%
\subsection{Text Graph Convolutional Networks (Text GCN)}
We build a large and heterogeneous text graph which contains word nodes and document nodes so that global word co-occurrence can be explicitly modeled and graph convolution can be easily adapted, as shown in Figure \ref{fig:framework}. The number of nodes in the text graph $|V|$ is the number of documents (corpus size) plus the number of unique words (vocabulary size) in a corpus. We simply set feature matrix $X = I$ as an identity matrix which means every word or document is represented as a one-hot vector as the input to Text GCN. We build edges among nodes based on word occurrence in documents (document-word edges) and word co-occurrence in the whole corpus (word-word edges).
%
%
The weight of the edge between a document node and a word node is the term frequency-inverse document frequency (TF-IDF) of the word in the document, where term frequency is the number of times the word appears in the document, inverse document frequency is the logarithmically scaled inverse fraction of the number of documents that contain the word. We found using TF-IDF weight is better than using term frequency only. To utilize global word co-occurrence information, we use a fixed size sliding window on all documents in the corpus to gather co-occurrence statistics. We employ point-wise mutual information (PMI), a popular measure for word associations, to calculate weights between two word nodes. We also found using PMI achieves better results than using word co-occurrence count in our preliminary experiments. Formally, the weight of edge between node $i$ and node $j$ is defined as
%
%
\begin{align}
A_{ij} = 
\begin{cases}
\quad\text{PMI}(i, j)& i,j \text{ are words, } \text{PMI}(i, j) > 0\\
\quad \text{TF-IDF}_{ij}& i \text{ is document, } j \text{ is word }\\
\quad 1 & i = j \\
\quad 0&  \text{otherwise}
\end{cases}
\end{align}
The PMI value of a word pair $i, j$ is computed as
\begin{equation}
\text{PMI}(i, j) = \log \frac{p(i, j)}{p(i)p(j)}
\end{equation}
\begin{equation}
p(i, j) = \frac{\#W(i, j)}{\#W}
\end{equation}
\begin{equation}
p(i) = \frac{\#W(i)}{\#W}
\end{equation}
where $\#W(i)$ is the number of sliding windows in a corpus that contain word $i$, $\#W(i, j)$ is the number of sliding windows that contain both word $i$ and $j$, and $\#W$ is the total number of sliding windows in the corpus. A positive PMI value implies a high semantic correlation of words in a corpus, while a negative PMI value indicates little or no semantic correlation in the corpus. Therefore, we only add edges between word pairs with positive PMI values.

After building the text graph, we feed the graph into a simple two layer GCN as in \cite{kipf2017semi}, the second layer node (word/document) embeddings have the same size as the labels set and are fed into a \textit{softmax} classifier:
\begin{equation}
Z = \text{softmax}(\tilde{A}\text{ ReLU}(\tilde{A}XW_0)W_1)
\label{2layer}
\end{equation}
where $\tilde{A} = D^{-\frac{1}{2}}A D^{-\frac{1}{2}}$ is the same as in equation \ref{1stl}, and $\text{softmax}(x_i) = \frac{1}{\mathcal{Z}} \exp(x_i)$ with $\mathcal{Z} = \sum_i \exp(x_i)$. The loss function is defined as the cross-entropy error over all labeled documents:
\begin{equation}
\mathcal{L} = -\sum_{d \in \mathcal{Y}_D} \sum_{f=1}^F Y_{df} \ln Z_{df}
\end{equation}
where $\mathcal{Y}_D$ is the set of document indices that have labels and $F$ is the dimension of the output features, which is equal to the number of classes. $Y$ is the label indicator matrix. The weight parameters $W_0$ and $W_1$ can be trained via gradient descent. In equation~\ref{2layer}, $E_1 = \tilde{A}XW_0$ contains the first layer document and word embeddings and $E_2 = \tilde{A}\text{ ReLU}(\tilde{A}XW_0)W_1$ contains the second layer document and word embeddings. The overall Text GCN model is schematically illustrated in Figure \ref{fig:framework}.
%
%

A two-layer GCN can allow message passing among nodes that are at maximum two steps away. Thus although there is no direct document-document edges in the graph, the two-layer GCN allows the information exchange between pairs of documents.
In our preliminary experiment. We found that a two-layer GCN performs better than a one-layer GCN, while more layers did not improve the performances. This is similar to results in~\cite{kipf2017semi} and \cite{DBLP:conf/aaai/LiHW18}.








\section{Experiment}
In this section we evaluate our Text Graph Convolutional Networks (Text GCN) on two experimental tasks. Specifically we want to determine:
\begin{itemize}
  \item Can our model achieve satisfactory results in text classification, even with limited labeled data?
  \item Can our model learn predictive word and document embeddings?
  \end{itemize}
\paragraph{Baselines.}
We compare our Text GCN with multiple state-of-the-art text classification and embedding methods as follows:
\begin{itemize}
\item \textbf{TF-IDF + LR} : bag-of-words model with term frequency-inverse document frequency weighting. Logistic Regression is used as the classifier. 
\item \textbf{CNN}: Convolutional Neural Network~\cite{kim2014convolutional}. We explored CNN-rand which uses randomly initialized word embeddings and CNN-non-static which uses pre-trained word embeddings.
\item \textbf{LSTM}: The LSTM model defined in~\cite{liu2016recurrent} which uses the last hidden state as  the representation of the whole text. We also experimented with the model with/without pre-trained word embeddings.
\item \textbf{Bi-LSTM}: a bi-directional LSTM, commonly used in text classification. We input pre-trained word embeddings to Bi-LSTM.
\item \textbf{PV-DBOW}: a paragraph vector model proposed by~\cite{le2014distributed}, the orders of words in text are ignored. We used Logistic Regression as the classifier.
\item \textbf{PV-DM}: a paragraph vector model proposed by~\cite{le2014distributed}, which considers the word order. We used Logistic Regression as the classifier.
\item \textbf{PTE}: predictive text embedding~\cite{tang2015pte}, which firstly learns word embedding based on heterogeneous text network containing words, documents and labels as nodes, then averages word embeddings as document embeddings for text classification. 
\item \textbf{fastText}: a simple and efficient text classification method~\cite{joulin2017bag}, which treats the average of word/n-grams embeddings as document embeddings, then feeds document embeddings into a linear classifier. We evaluated it with and without bigrams. 
\item \textbf{SWEM}: simple word embedding models~\cite{Shen2018Baseline}, which employs simple pooling strategies operated over word embeddings.
\item \textbf{LEAM}: label-embedding attentive models~\cite{P18-1216}, which embeds the words and labels in the same joint space for text classification. It utilizes label descriptions.
%
%
\item \textbf{Graph-CNN-C}: a graph CNN model that operates convolutions over word embedding similarity graphs~\cite{defferrard2016convolutional}, in which Chebyshev filter is used.
\item \textbf{Graph-CNN-S}: the same as Graph-CNN-C but using Spline filter~\cite{bruna2013spectral}.
\item \textbf{Graph-CNN-F}: the same as Graph-CNN-C but using Fourier filter~\cite{henaff2015deep}.
\end{itemize}

    {\small
    \begin{table*}[t]\footnotesize
    \centering
    \renewcommand{\arraystretch}{1.2}
    \caption{Summary statistics of datasets.}
    \begin{tabular}{c|ccccccc}
    \hline
    \bf{Dataset}& \bf{\# Docs}	& \bf{\# Training}& \bf{\# Test}& \bf{\# Words} & \bf{\# Nodes}& \bf{\# Classes} & \bf{Average Length} \\
    \hline
    20NG & 18,846 & 11,314 & 7,532 & 42,757 & 61,603 & 20  & 221.26\\
   
     R8 & 7,674 &5,485 & 2,189 & 7,688 & 15,362 & 8 & 65.72\\
    
     R52& 9,100 & 6,532	 & 2,568 & 8,892& 17,992 & 52 & 69.82\\
    
    Ohsumed& 7,400 & 3,357 & 4,043 & 14,157 & 21,557 & 23 & 135.82\\
    MR& 10,662 & 7,108 & 3,554 & 18,764 & 29,426 & 2 &20.39\\
    \hline
    \end{tabular}
    \label{tab:statistics}
    \end{table*}
    }
    
\paragraph{Datasets.}
We ran our experiments on five widely used benchmark corpora including 20-Newsgroups (20NG),  Ohsumed, R52 and R8 of Reuters 21578 and Movie Review (MR). 
\begin{itemize}
\item The 20NG dataset\footnote{http://qwone.com/\url{~}jason/20Newsgroups/} (“bydate” version) contains 18,846 documents evenly categorized into 20 different categories. In total, 11,314 documents are in the training set and 7,532 documents are in the test set.
\item The Ohsumed corpus\footnote{http://disi.unitn.it/moschitti/corpora.htm} is from the MEDLINE database, which is a bibliographic database of important medical literature maintained by the National Library of Medicine. In this work, we used the 13,929 unique cardiovascular diseases abstracts in the first 20,000 abstracts of the year 1991.  Each document in the set has one or more associated categories from the 23 disease categories. As we focus on single-label text classification, the documents belonging to multiple categories are excluded so that 7,400 documents belonging to only one category remain. 3,357 documents are in the training set and 4,043 documents are in the test set. 
\item R52 and R8\footnote{https://www.cs.umb.edu/\url{~}smimarog/textmining/datasets/} (all-terms version) are two subsets of the Reuters 21578 dataset. R8 has 8 categories, and was split to 5,485 training and 2,189 test documents. R52 has 52 categories, and was split to 6,532 training and 2,568 test documents. 
\item MR is a movie review dataset for binary sentiment classification, in which each review only contains one sentence ~\cite{pang2005seeing}\footnote{http://www.cs.cornell.edu/people/pabo/movie-review-data/}. The corpus has 5,331 positive and 5,331 negative reviews. We used the training/test split in~\cite{tang2015pte}\footnote{https://github.com/mnqu/PTE/tree/master/data/mr}.
\end{itemize}
We first preprocessed all the datasets by cleaning and tokenizing text as~\cite{kim2014convolutional}. We then removed stop words defined in NLTK\footnote{http://www.nltk.org/} and low frequency words appearing less than 5 times for 20NG, R8, R52 and Ohsumed. The only exception was MR, we did not remove words after cleaning and tokenizing raw text, as the documents are very short. The statistics of the preprocessed datasets are summarized in Table \ref{tab:statistics}.

    \begin{table*}[t]\footnotesize
    \centering
    \renewcommand{\arraystretch}{1.2}
    
    \caption{Test Accuracy on document classification task. We run all models 10 times and report mean $\pm$ standard deviation. Text GCN significantly outperforms baselines on 20NG, R8, R52 and Ohsumed based on student $t$-test ($p <0.05$).}
    \begin{tabular}{c|c|c|c|c|c}
    \hline
    \bf{Model}& \bf{20NG}	& \bf{R8}& \bf{R52}& \bf{Ohsumed} & \bf{MR}\\
    \hline
 TF-IDF + LR & 0.8319 $\pm$ 0.0000 & 0.9374 $\pm$ 0.0000 & 0.8695 $\pm$ 0.0000 & 0.5466 $\pm$ 0.0000 & 0.7459 $\pm$ 0.0000 \\
   
     CNN-rand &0.7693 $\pm$ 0.0061&0.9402 $\pm$ 0.0057 & 0.8537 $\pm$ 0.0047 & 0.4387 $\pm$ 0.0100& 0.7498 $\pm$ 0.0070 \\
     CNN-non-static &0.8215 $\pm$ 0.0052& 0.9571 $\pm$ 0.0052& 0.8759 $\pm$ 0.0048 & 0.5844 $\pm$ 0.0106& \bf{0.7775 $\pm$ 0.0072}  \\
    LSTM &0.6571 $\pm$ 0.0152& 0.9368 $\pm$ 0.0082& 0.8554 $\pm$ 0.0113 &0.4113 $\pm$ 0.0117 & 0.7506 $\pm$ 0.0044 \\
    LSTM (pretrain) &0.7543 $\pm$ 0.0172& 0.9609 $\pm$ 0.0019&0.9048 $\pm$ 0.0086 &0.5110 $\pm$ 0.0150 & 0.7733 $\pm$ 0.0089 \\
    Bi-LSTM &0.7318 $\pm$ 0.0185& 0.9631 $\pm$ 0.0033& 0.9054 $\pm$ 0.0091 &0.4927 $\pm$ 0.0107& 0.7768 $\pm$ 0.0086 \\
    PV-DBOW &0.7436 $\pm$ 0.0018& 0.8587 $\pm$ 0.0010& 0.7829 $\pm$ 0.0011 & 0.4665 $\pm$ 0.0019& 0.6109 $\pm$ 0.0010  \\
    PV-DM &0.5114 $\pm$ 0.0022& 0.5207 $\pm$ 0.0004&  0.4492 $\pm$ 0.0005& 0.2950 $\pm$ 0.0007& 0.5947 $\pm$ 0.0038 \\
    PTE & 0.7674 $\pm$ 0.0029&0.9669 $\pm$ 0.0013 & 0.9071 $\pm$ 0.0014 & 0.5358 $\pm$ 0.0029 & 0.7023 $\pm$ 0.0036\\
    fastText &0.7938 $\pm$ 0.0030& 0.9613 $\pm$ 0.0021& 0.9281 $\pm$  0.0009  & 0.5770 $\pm$ 0.0049 &  0.7514 $\pm$ 0.0020\\
    fastText (bigrams) & 0.7967 $\pm$ 0.0029& 0.9474 $\pm$ 0.0011& 0.9099 $\pm$ 0.0005 &0.5569 $\pm$ 0.0039 & 0.7624 $\pm$ 0.0012 \\
    SWEM &0.8516 $\pm$ 0.0029& 0.9532 $\pm$ 0.0026& 0.9294 $\pm$ 0.0024 & 0.6312 $\pm$ 0.0055& 0.7665 $\pm$ 0.0063 \\
    LEAM &0.8191 $\pm$ 0.0024& 0.9331 $\pm$ 0.0024 & 0.9184 $\pm$ 0.0023 & 0.5858 $\pm$ 0.0079& 0.7695 $\pm$ 0.0045 \\
    Graph-CNN-C &0.8142 $\pm$ 0.0032& 0.9699 $\pm$ 0.0012&  0.9275 $\pm$ 0.0022& 0.6386 $\pm$ 0.0053&  0.7722 $\pm$  0.0027\\
    Graph-CNN-S &--& 0.9680 $\pm$ 0.0020& 0.9274 $\pm$ 0.0024 &0.6282 $\pm$ 0.0037 & 0.7699 $\pm$ 0.0014 \\
	Graph-CNN-F &--&0.9689 $\pm$ 0.0006 & 0.9320 $\pm$ 0.0004 &0.6304 $\pm$ 0.0077& 0.7674 $\pm$ 0.0021 \\
   	Text GCN &\bf{0.8634 $\pm$ 0.0009}& \bf{0.9707 $\pm$ 0.0010}& \bf{0.9356 $\pm$ 0.0018}& \bf{0.6836 $\pm$ 0.0056}& 0.7674 $\pm$ 0.0020 \\
    \hline
    \end{tabular}
    \label{tab:results}
    \end{table*}

\paragraph{Settings.}
For Text GCN, we set the embedding size of the first convolution layer as 200 and set the window size as 20. We also experimented with other settings and found that small changes did not change the results much. We tuned other parameters and set the learning rate as 0.02, dropout rate as 0.5, $L_2$ loss weight as 0. We randomly selected $10\%$ of training set as validation set. Following \cite{kipf2017semi}, we trained Text GCN for a maximum of 200 epochs using Adam \cite{kinga2015method} and stop training if the validation loss does not decrease for 10 consecutive epochs. 
For baseline models, we used default parameter settings as in their original papers or implementations. For baseline models using pre-trained word embeddings, we used 300-dimensional GloVe word embeddings~\cite{pennington2014glove}\footnote{http://nlp.stanford.edu/data/glove.6B.zip}.

\paragraph{Test Performance.}
Table \ref{tab:results} presents test accuracy of each model. Text GCN performs the best and significantly outperforms all baseline models ($p < 0.05$ based on student $t$-test) on four datasets, which showcases the effectiveness of the proposed method on long text datasets. For more in-depth performance analysis, we note that TF-IDF + LR performs well on long text datasets like 20NG and can outperform CNN with randomly initialized word embeddings. When pre-trained GloVe word embeddings are provided, CNN performs much better, especially on Ohsumed and 20NG. CNN also achieves the best results on short text dataset MR with pre-trained word embeddings, which shows it can model consecutive and short-distance semantics well. Similarly, LSTM-based models also rely on pre-trained word embeddings and tend to perform better when documents are shorter. PV-DBOW achieves comparable results to strong baselines on 20NG and Ohsumed, but the results on shorter text are clearly inferior to others. This is likely due to the fact that word orders are important in short text or sentiment classification. PV-DM performs worse than PV-DBOW, the only comparable results are on MR, where word orders are more essential. The results of PV-DBOW and PV-DM indicate that unsupervised document embeddings are not very discriminative in text classification. PTE and fastText clearly outperform PV-DBOW and PV-DM because they learn document embeddings in a supervised manner so that label information can be utilized to learn more discriminative embeddings. The two recent methods SWEM and LEAM perform quite well, which demonstrates the effectiveness of simple pooling methods and label descriptions/embeddings. Graph-CNN models also show competitive performances. This suggests that building word similarity graph using pre-trained word embeddings can preserve syntactic and semantic relations among words, which can provide additional information in large external text data. 

The main reasons why Text GCN works well are two fold: 1) the text graph can capture both document-word relations and global word-word relations; 2) the GCN model, as a special form of Laplacian smoothing, computes the new features of a node as the weighted average of itself and its second order neighbors~\cite{DBLP:conf/aaai/LiHW18}. The label information of document nodes can be passed to their neighboring word nodes (words within the documents), then relayed to other word nodes and document nodes that are neighbor to the first step neighboring word nodes. Word nodes can gather comprehensive document label information and act as bridges or key paths in the graph, so that label information can be propagated to the entire graph. However, we also observed that Text GCN did not outperform CNN and LSTM-based models on MR. This is because GCN ignores word orders that are very useful in sentiment classification, while CNN and LSTM model consecutive word sequences explicitly. Another reason is that the edges in MR text graph are fewer than other text graphs, which limits the message passing among the nodes. There are only few document-word edges because the documents are very short. The number of word-word edges is also limited due to the small number of sliding windows. Nevertheless, CNN and LSTM rely on pre-trained word embeddings from external corpora while Text GCN only uses information in the target input corpus.
%
%

\begin{figure}[!htb]
\centering
\subfigure[R8]{
\label{fig:swin:a} %
\includegraphics[height = 29 mm]{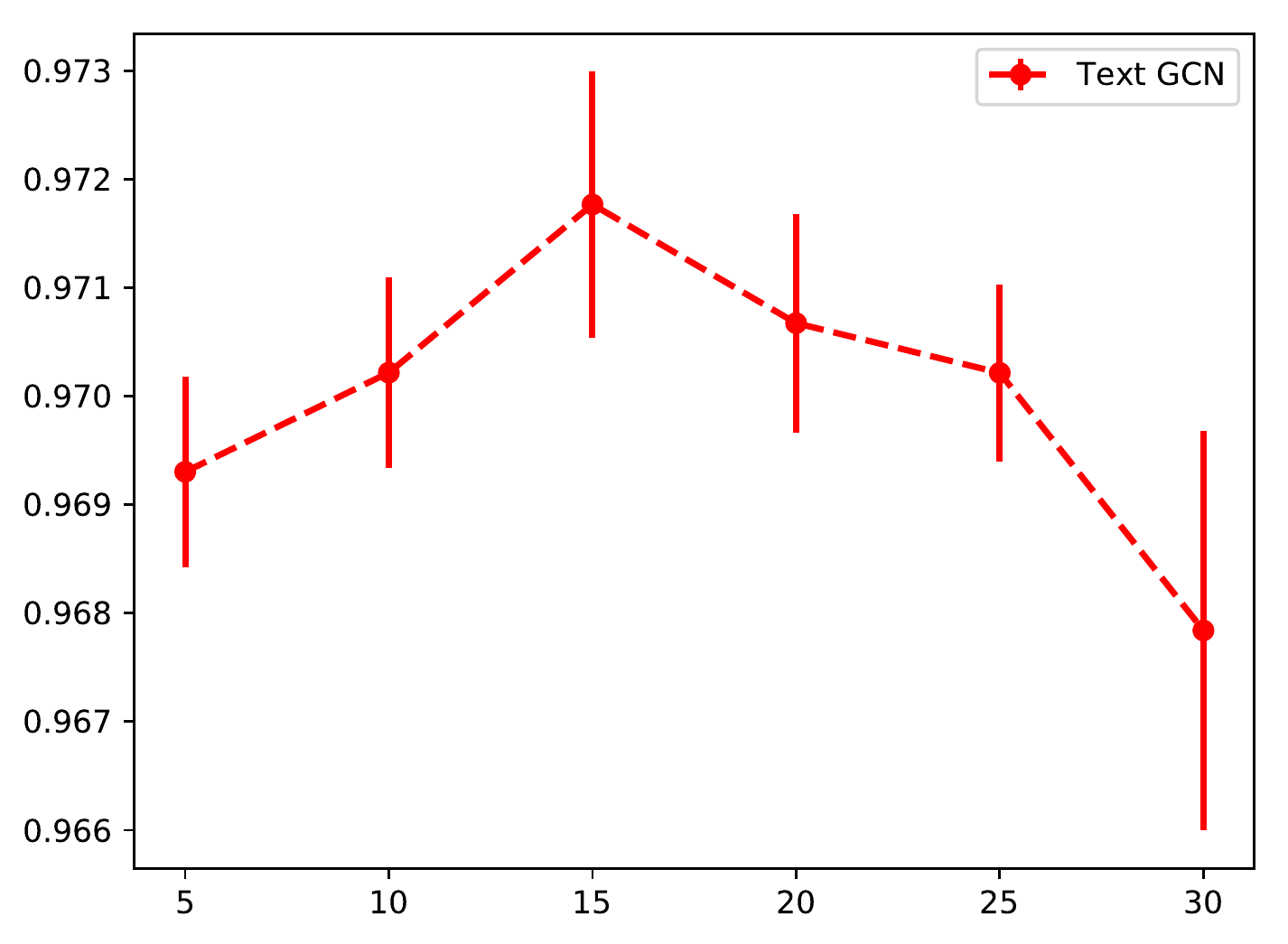}}
\subfigure[MR]{
\label{fig:swin:b} %
\includegraphics[height = 29 mm]{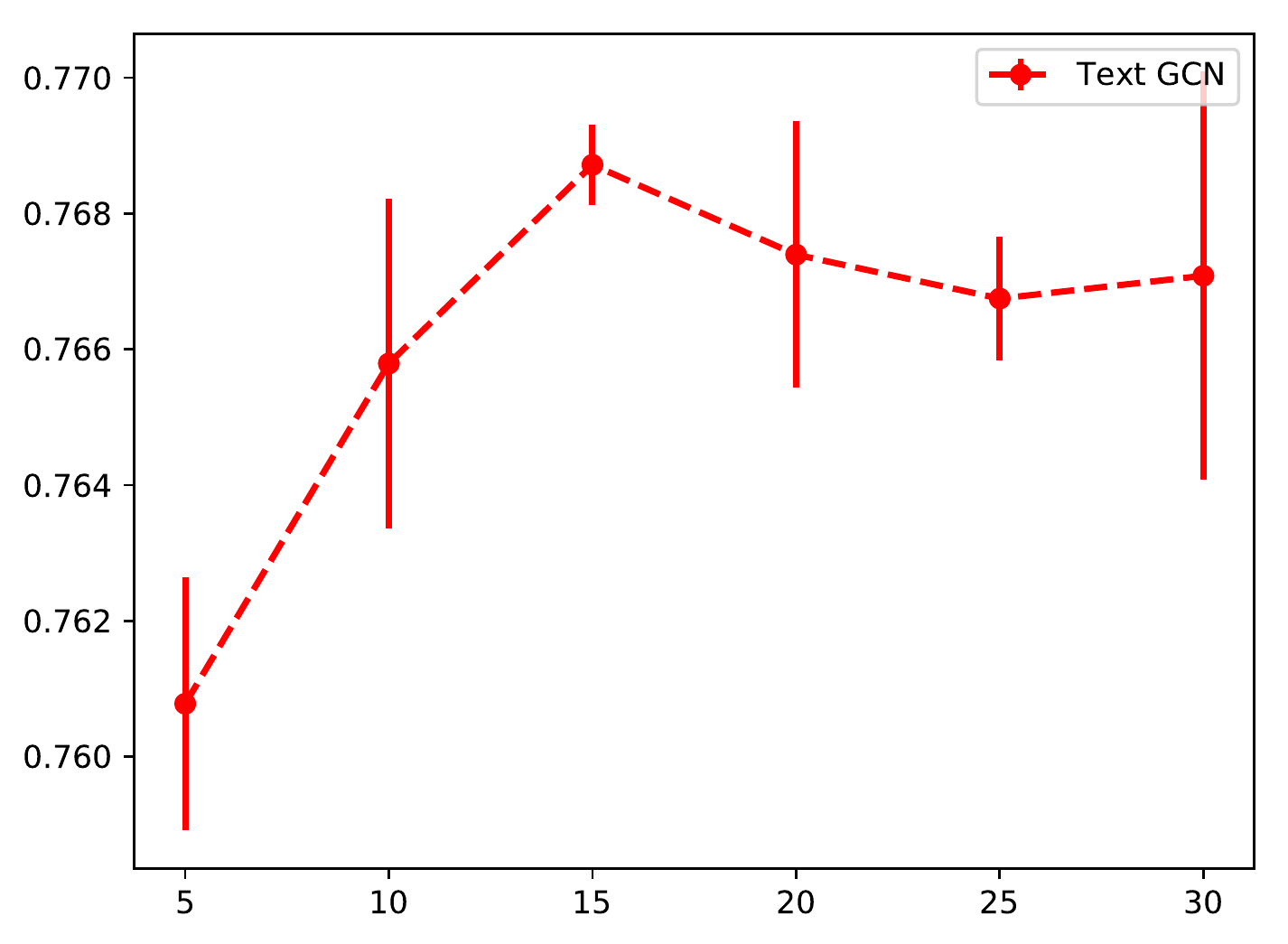}}
\caption{Test accuracy with different sliding window sizes.}
\label{fig:swin}
\end{figure}

\begin{figure}[!htb]
\centering
\subfigure[R8]{
\label{fig:edim:a} %
\includegraphics[height = 29 mm]{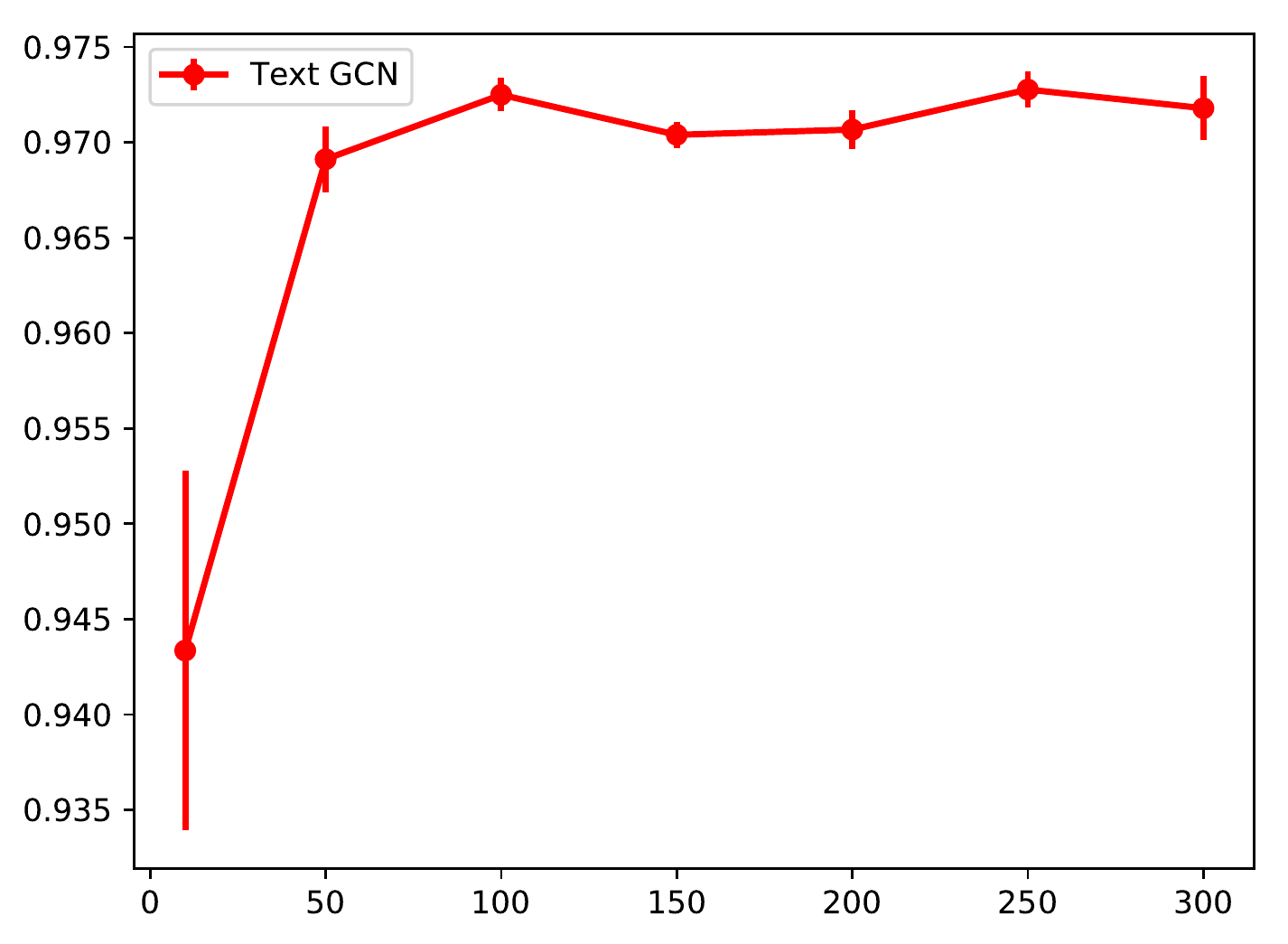}}
\subfigure[MR]{
\label{fig:edim:b} %
\includegraphics[height = 29 mm]{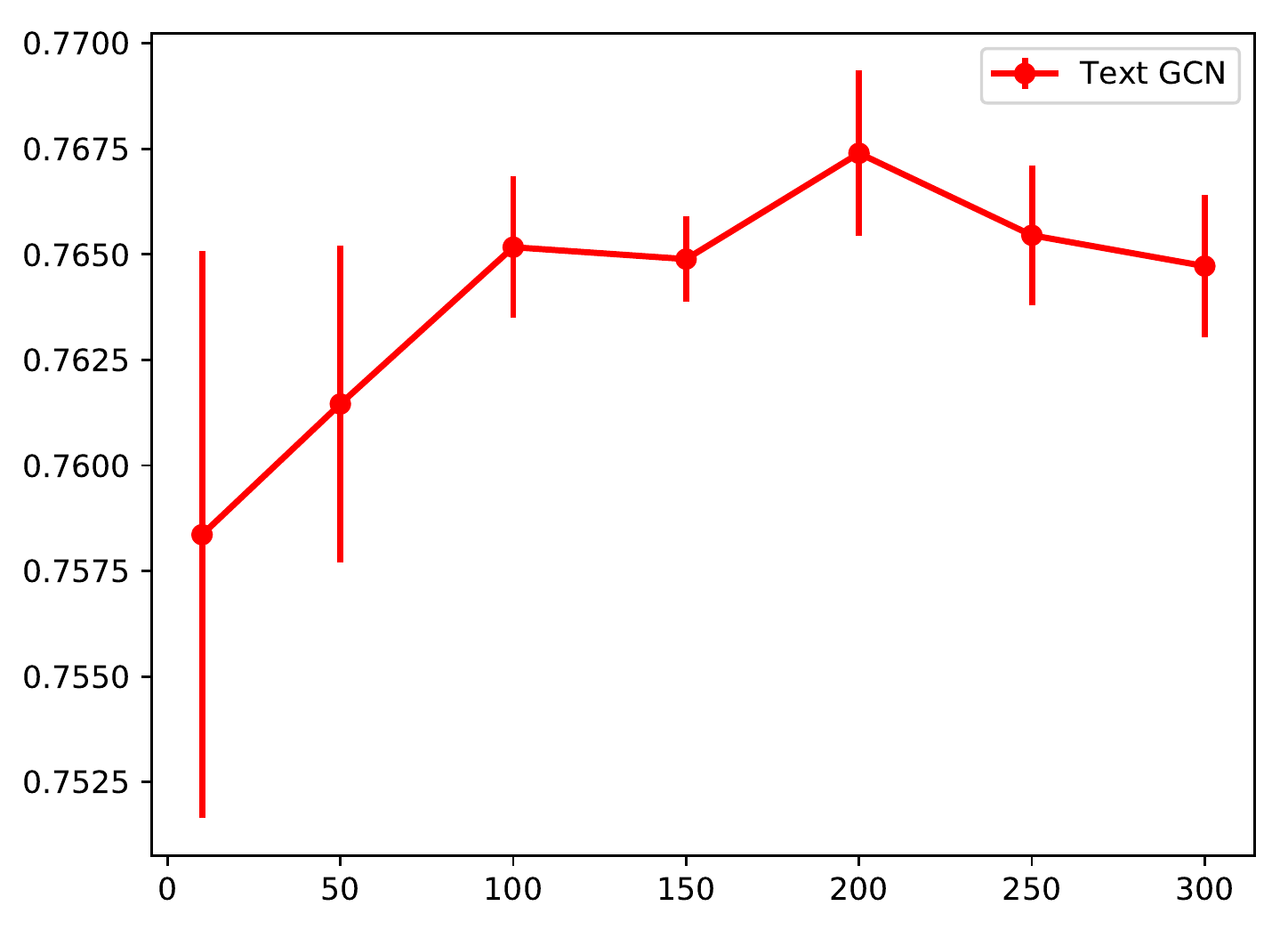}}
\caption{Test accuracy by varying embedding dimensions.}
\label{fig:edim} %
\end{figure}

\begin{figure}[t]
\centering
\subfigure[20NG]{
\label{fig:proportion:a} %
\includegraphics[height = 30.5 mm]{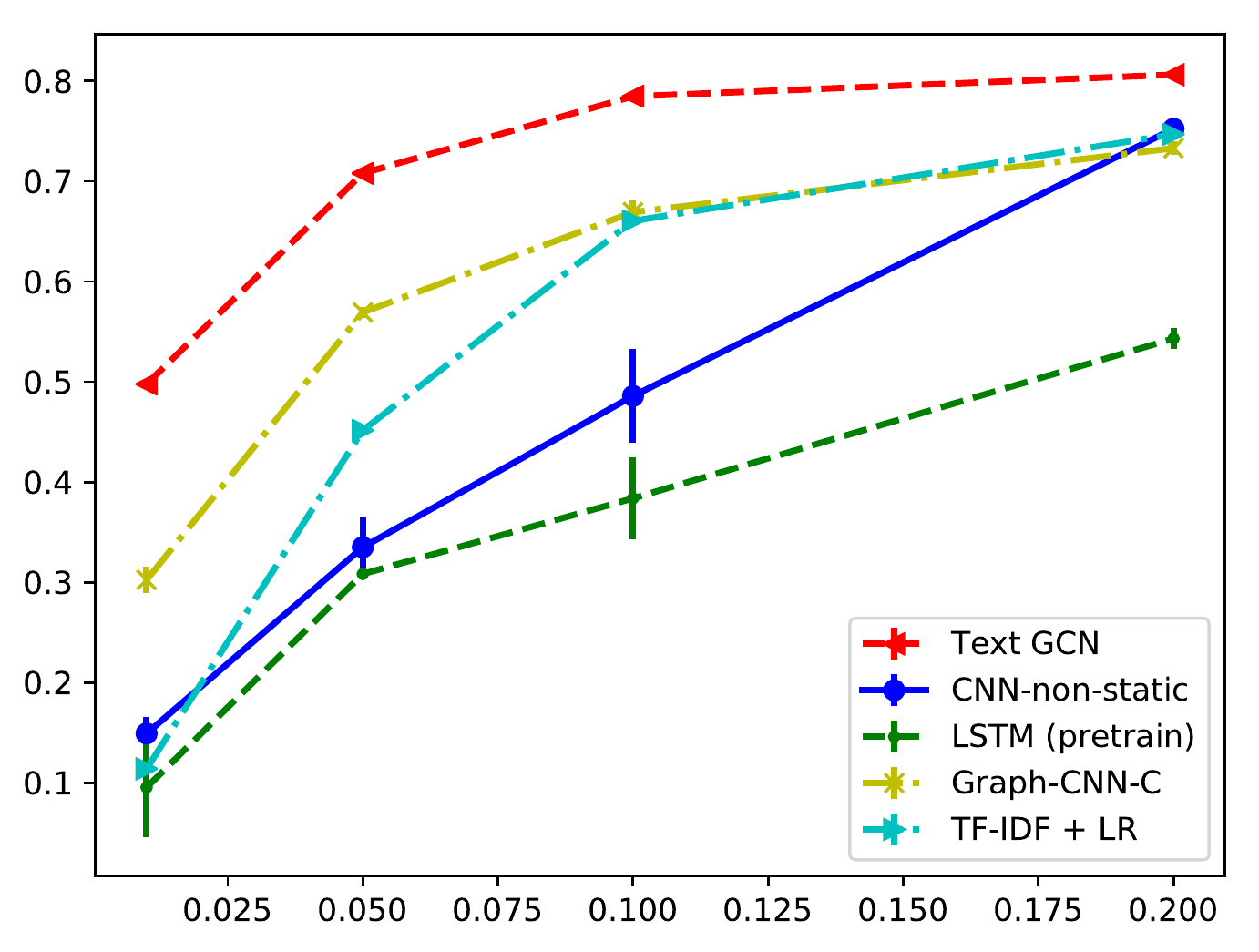}}
\subfigure[R8]{
\label{fig:proportion:b} %
\includegraphics[height = 30.5 mm]{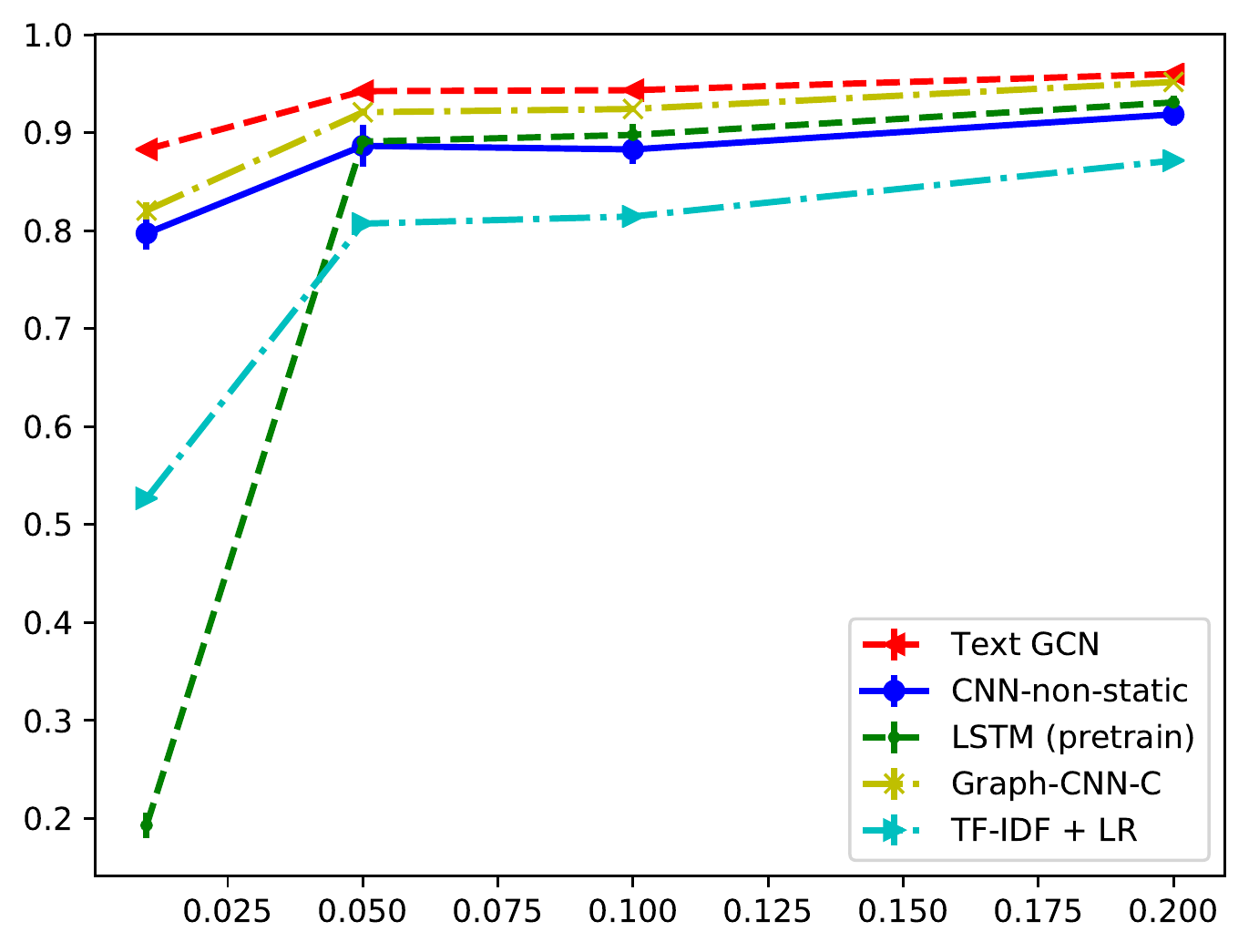}}
\caption{Test accuracy by varying training data proportions.}
\label{fig:proportion}
\end{figure}

\begin{figure}[t]
\centering
\subfigure[Text GCN, 1st layer]{
\label{fig:doc2vec:a} %
\includegraphics[height = 25 mm]{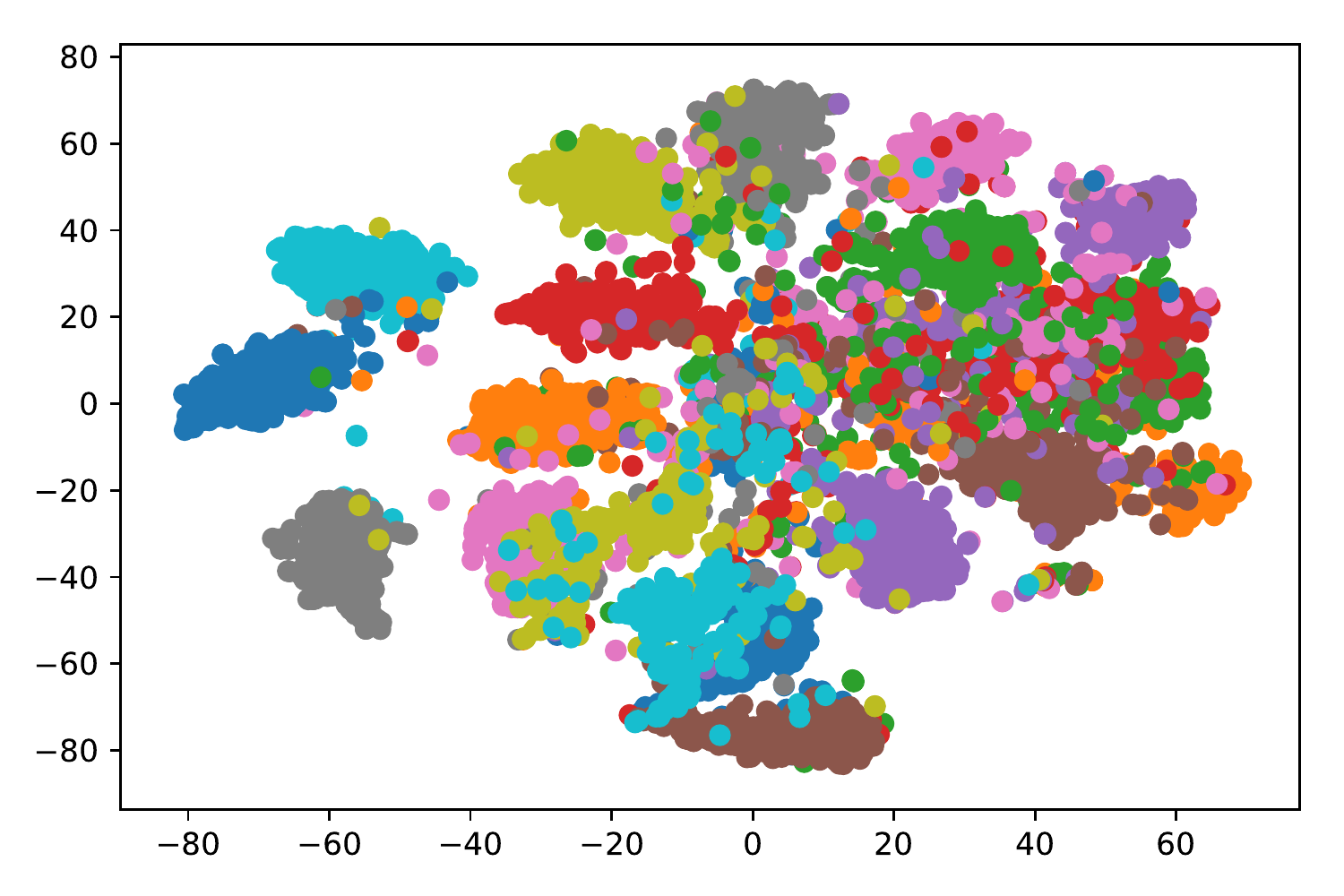}}
\subfigure[Text GCN, 2nd layer]{
\label{fig:doc2vec:b} %
\includegraphics[height = 25 mm]{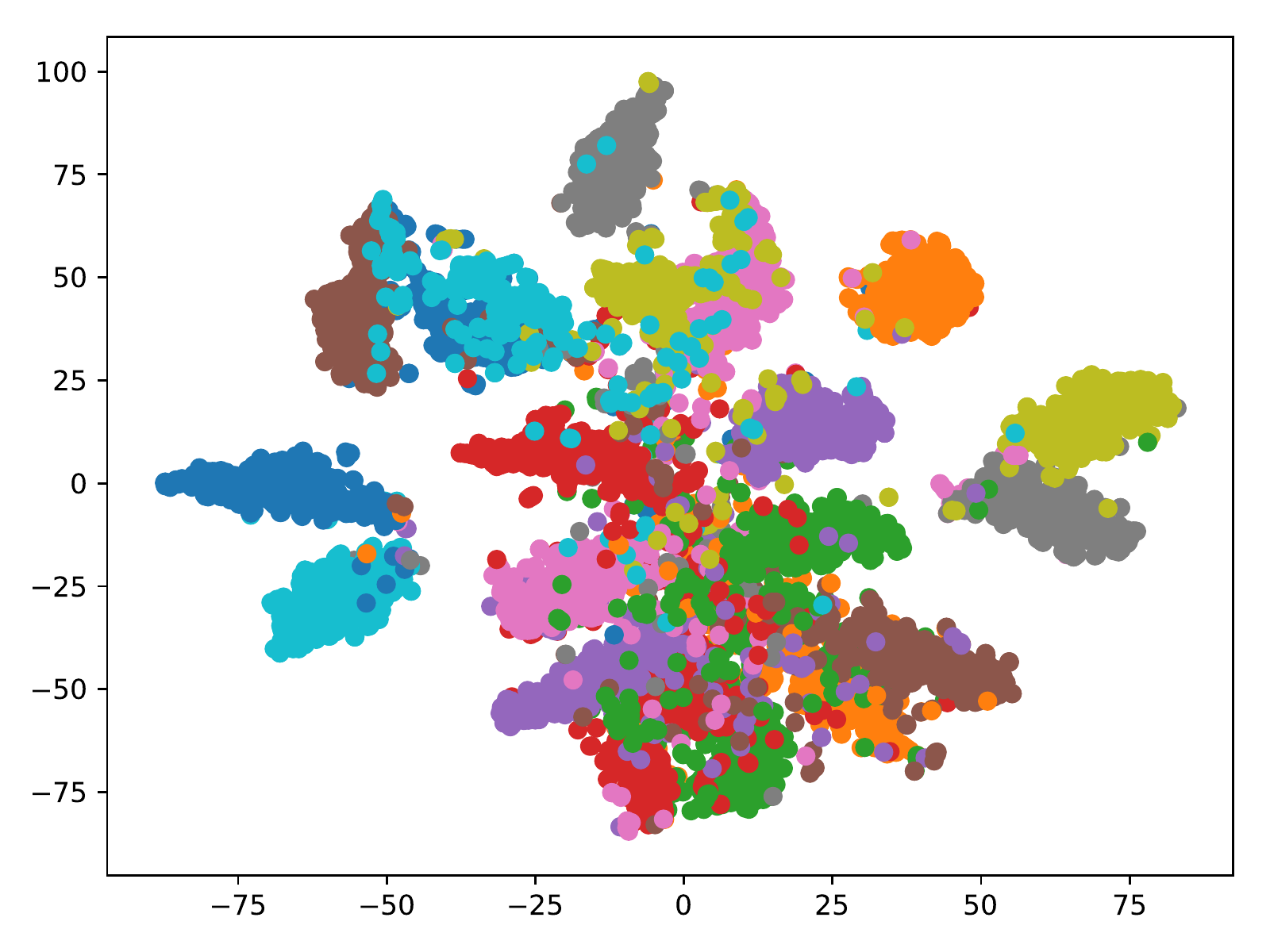}}

\subfigure[PV-DBOW]{
\label{fig:doc2vec:c} %
\includegraphics[height = 25 mm]{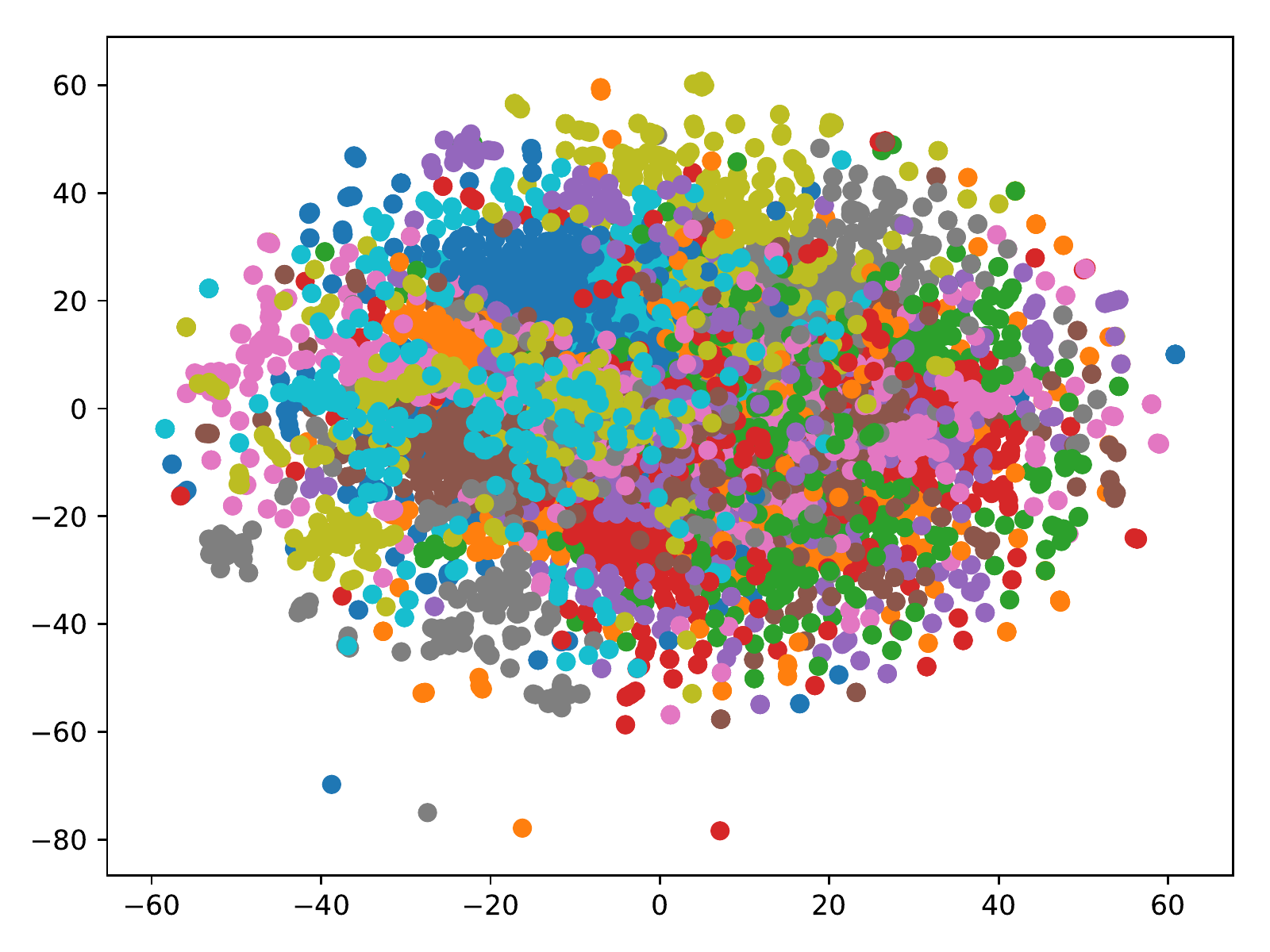}}
\subfigure[PTE]{
\label{fig:doc2vec:d} %
\includegraphics[height = 25 mm]{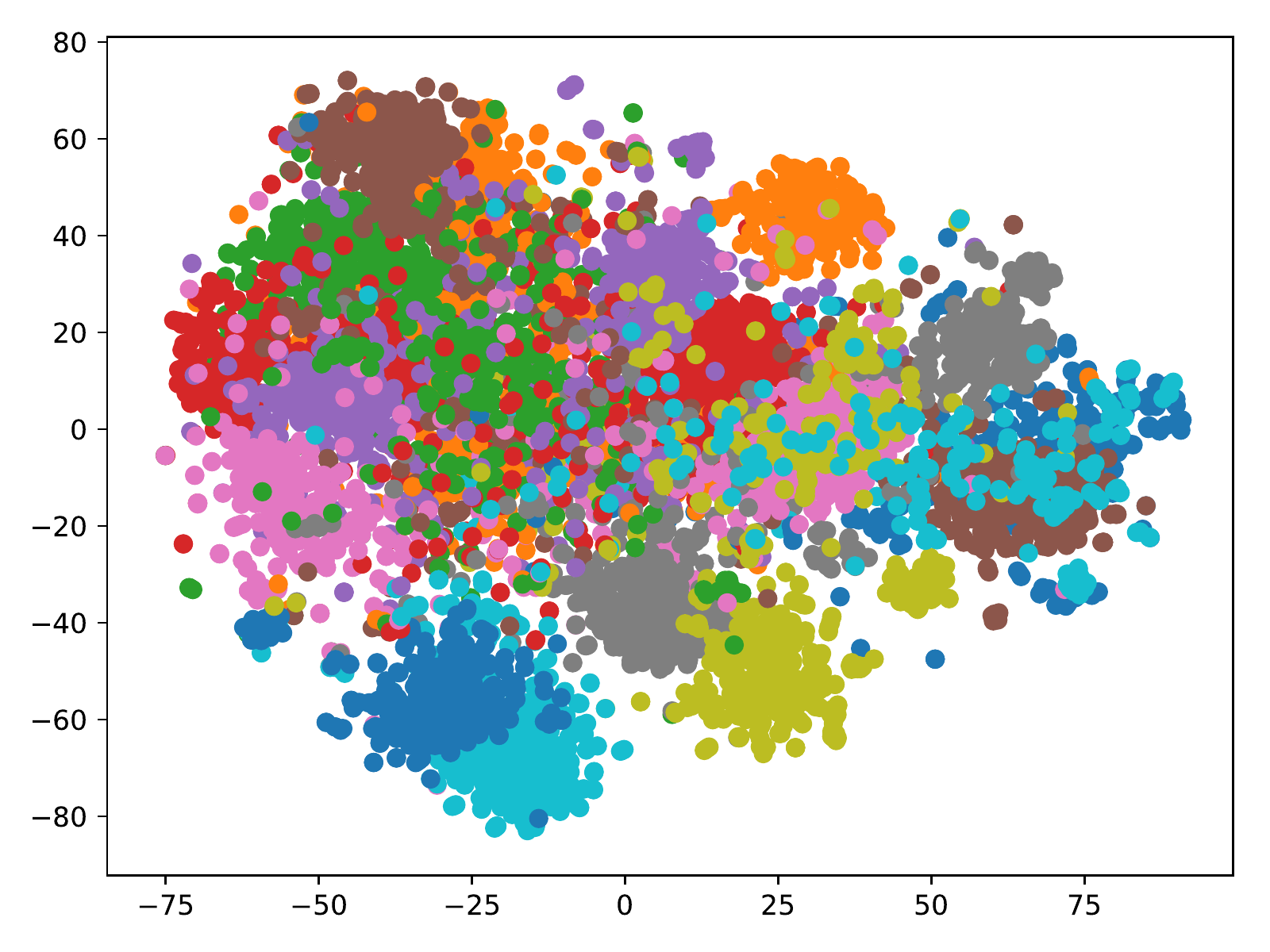}}

\caption{The t-SNE visualization of test set document embeddings in 20NG.}
\label{fig:doc2vec}
\end{figure}

\paragraph{Parameter Sensitivity.} Figure \ref{fig:swin} shows test accuracies with different sliding window sizes on R8 and MR. We can see that test accuracy first increases as window size becomes larger, but the average accuracy stops increasing when window size is larger than 15. This suggests that too small window sizes could not generate sufficient global word co-occurrence information, while too large window sizes may add edges between nodes that are not very closely related. Figure \ref{fig:edim} depicts the classification performance on R8 and MR with different dimensions of the-first layer embeddings. We observed similar trends as in Figure \ref{fig:swin}. Too low dimensional embeddings may not propagate label information to the whole graph well, while high dimensional embeddings do not improve classification performances and may cost more training time.


\paragraph{Effects of the Size of Labeled Data.} 
 
In order to evaluate the effect of the size of the labeled data, we tested several best performing models with different proportions of the training data. Figure \ref{fig:proportion} reports test accuracies with $1\%$, $5\%$, $10\%$ and $20\%$ of original 20NG and R8 training set. We note that Text GCN can achieve higher test accuracy with limited labeled documents. For instance, Text GCN achieves a test accuracy of 0.8063 $\pm$ 0.0025 on 20NG with only $20\%$ training documents and a test accuracy of 0.8830 $\pm$ 0.0027 on R8 with only $1\%$ training documents which are higher than some baseline models with even the full training documents. These encouraging results are similar to results in \cite{kipf2017semi} where GCN can perform quite well with low label rate, which again suggests that GCN can propagate document label information to the entire graph well and our word document graph preserves global word co-occurrence information.

\paragraph{Document Visualization.}
We give an illustrative visualization of the document embeddings leaned by Text GCN. We use t-SNE tool~\cite{maaten2008visualizing} to visualize the learned document embeddings. Figure~\ref{fig:doc2vec} shows the visualization of 200 dimensional 20NG test document embeddings learned by GCN (first layer), PV-DBOW and PTE. We also show 20 dimensional second layer test document embeddings of Text GCN. We observe that Text GCN can learn more discriminative document embeddings, and the second layer embeddings are more distinguishable than the first layer.

\paragraph{Word Visualization.}

\begin{figure}[t]
\centering
\includegraphics[height = 30 mm]{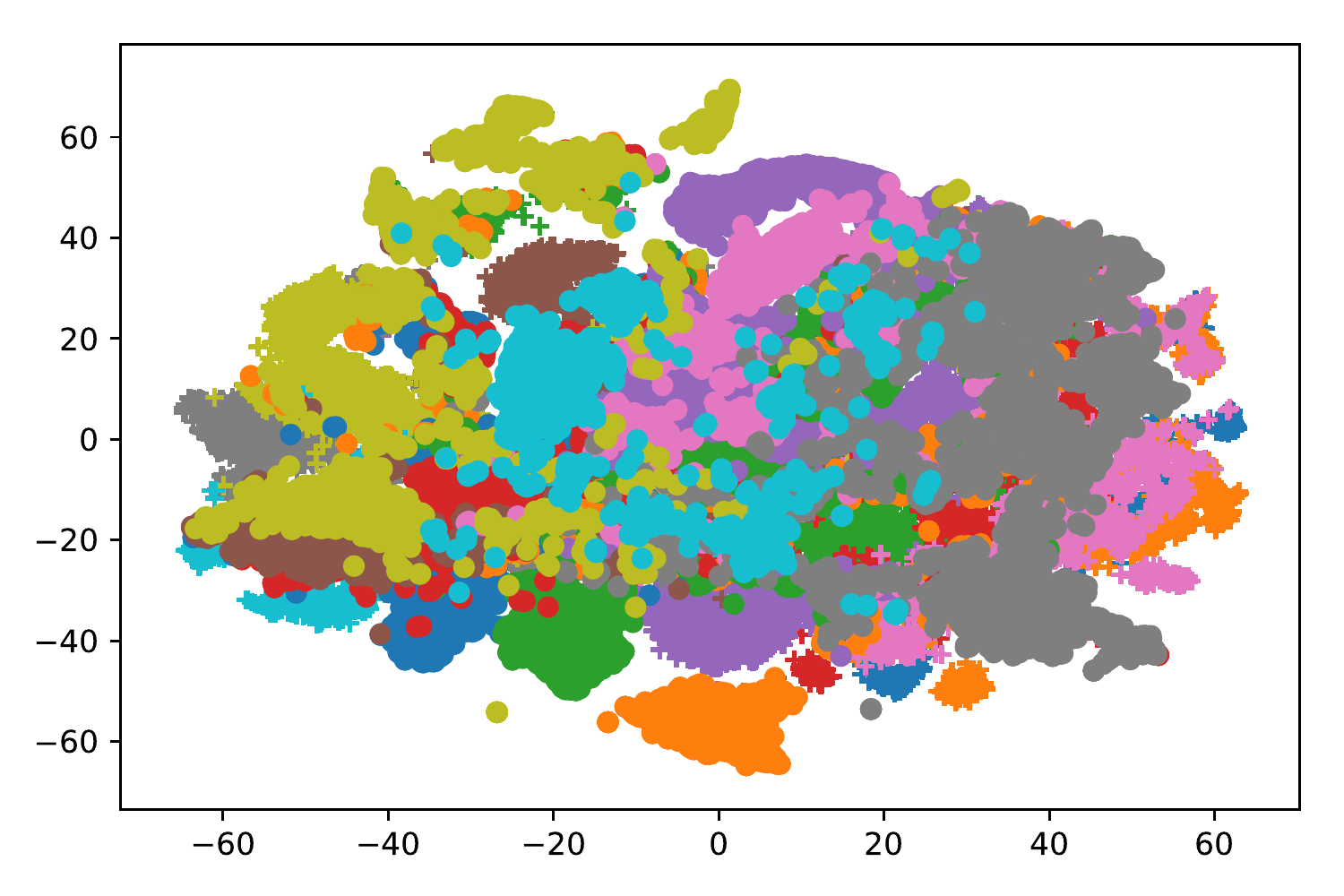}
\caption{The t-SNE visualization of the second layer word embeddings (20 dimensional) learned from 20NG. We set the dimension with the largest value as a word's label.}
\label{fig:word2vec}
\end{figure}

We also qualitatively visualize word embeddings learned by Text GCN.
Figure~\ref{fig:word2vec} shows the t-SNE visualization of the second layer word embeddings learned from 20NG. We set the dimension with the highest value as a word's label. We can see that words with the same label are close to each other, which means most words are closely related to some certain document classes. We also show top 10 words with highest values under each class in Table 3. We note that the top 10 words are interpretable. For example, ``jpeg'', ``graphics'' and ``image'' in column 1 can represent the meaning of their label ``comp.graphics'' well. Words in other columns can also indicate their label's meaning.

  \begin{table}[t]\footnotesize
  \centering
  \renewcommand{\arraystretch}{1.2}
  \label{table_1}
  \caption{Words with highest values for several classes in 20NG. Second layer word embeddings are used. We show top 10 words for each class.}
  \begin{tabular}{c|c|c|c}
  \hline
  
   comp.graphics & sci.space &sci.med & rec.autos \\
  \hline
  jpeg& space & candida  & car\\
  
  graphics & orbit & geb & cars\\
  
  image  &  shuttle & disease & v12\\
  
  gif  & launch & patients & callison\\
  
  3d &  moon & yeast & engine\\
  
  images &  prb & msg & toyota\\
 
  rayshade &  spacecraft & vitamin & nissan \\
  
  polygon & solar & syndrome & v8 \\
  
  pov & mission & infection & mustang\\
  
  viewer & alaska & gordon & eliot \\
  \hline
  \end{tabular}
  \end{table}

\paragraph{Discussion.}
From experimental results, we can see the proposed Text GCN can achieve strong text classification results and learn predictive document and word embeddings. However, a major limitation of this study is that the GCN model is inherently transductive, in which test document nodes (without labels) are included in GCN training. Thus Text GCN could not quickly generate embeddings and make prediction for unseen test documents. Possible solutions to the problem are introducing inductive~\cite{hamilton2017inductive} or fast GCN model~\cite{chen2018fastgcn}.
%
%

\section{Conclusion and Future Work}
In this study, we propose a novel text classification method termed Text Graph Convolutional Networks (Text GCN). We build a heterogeneous word document graph for a whole corpus and turn document classification into a node classification problem. Text GCN can capture global word co-occurrence information and utilize limited labeled documents well. A simple two-layer Text GCN demonstrates promising results by outperforming numerous state-of-the-art methods on multiple benchmark datasets.
%
%

In addition to generalizing Text GCN model to inductive settings, some interesting future directions include improving the classification performance using attention mechanisms~\cite{velivckovic2017graph} and developing unsupervised text GCN framework for representation learning on large-scale unlabeled text data. 

\section*{Acknowledgments}

This work is supported in part by NIH grant R21LM012618.

\bibliographystyle{aaai}
\bibliography{aaai19}

\end{document}